\documentclass[lettersize,journal]{IEEEtran}
\usepackage{amsmath,amsfonts}
\usepackage{algorithmic}
\usepackage{algorithm}
\usepackage{array}
\usepackage[caption=false,font=normalsize,labelfont=sf,textfont=sf]{subfig}
\usepackage{textcomp}
\usepackage{stfloats}
\usepackage{placeins}
\usepackage{url}
\usepackage{verbatim}
\usepackage{graphicx}
\usepackage{cite}
\usepackage{listings}

\usepackage{booktabs}
\PassOptionsToPackage{table}{xcolor}
\usepackage{tikz}
\usepackage{pgfplots}
\pgfplotsset{compat=1.18}
\usepackage{booktabs}
\usepackage{amssymb}
\usepackage{graphicx}
\usepackage{makecell}
\usepackage{color}
\usepackage{booktabs}
\usepackage{xcolor}
\usepackage{pifont}
\usepackage{makecell}
\usepackage{array}
\usepackage{amsthm}   
\usepackage{amsmath}   
\usepackage{amsthm} 

\definecolor{lightgray}{rgb}{0.95,0.95,0.95}
\definecolor{darkblue}{rgb}{0.0,0.0,0.6}
\definecolor{cyan}{rgb}{0.0,0.6,0.6}
\begin{document}

\title{Null-Space Constrained Low-Rank Adaptation for Response-Specified Large Language Model Unlearning}

\author{Bocheng Ju, Jianhua~Wang, Chengliang~Liu, and Xiaolin~Chang
\IEEEcompsocitemizethanks{
\IEEEcompsocthanksitem Bocheng Ju and Xiaolin Chang are with the Beijing Key Laboratory of Security and Privacy in Intelligent Transportation, Beijing Jiaotong University, P.R.China. (e-mail: xlchang@bjtu.edu.cn)
\IEEEcompsocthanksitem Jianhua Wang is with the College of Computer Science and Technology, Taiyuan University of Technology, Taiyuan, China, 030024. (e-mail: wangjianhua02@tyut.edu.cn)
\IEEEcompsocthanksitem Chengliang Liu is with the Institute of Computing Technologies, China Academy of Railway Sciences Corporation Limited, Beijing 100081, China. (e-mail: liucl@rails.cn).

}}



\maketitle

\begin{abstract}
Large language model unlearning aims to suppress designated undesirable knowledge while preserving benign capabilities.
Many unlearning objectives focus on suppressing undesired answers, while recent target-guided variants specify replacement behavior but still leave update locality largely unconstrained.
This paper introduces \emph{Null-Space Constrained Response-Specified Unlearning} (NSRU), a projection-constrained low-rank framework for controlled LLM unlearning.
NSRU uses an explicitly structured safe target response to specify the desired behavior for each forget query, while suppressing the original undesired content.
To localize adaptation, NSRU estimates per-module retain subspaces from benign hidden representations and uses an orthogonal-projected low-rank parameterization to confine LoRA updates to the null space of the retain subspace.
The resulting objective jointly optimizes safe-target learning, undesired-response suppression, and retention preservation under this constrained parameterization.
We provide a local first-order analysis showing that the projected update reduces retain-side perturbations while preserving editable directions for shaping forget-query behavior.
Experiments on TOFU show that NSRU effectively suppresses extractable forget-set knowledge while improving retain QA performance, model utility, and safe-target alignment over representative baselines.
On WMDP, NSRU keeps hazardous-domain accuracy near the random-choice region while preserving broad and domain-adjacent MMLU utility.
Ablation studies support the complementary roles of safe-target supervision, undesired-response suppression, retention loss, and null-space projected updates, while sensitivity and robustness analyses indicate stable behavior across the tested hyperparameter and prompt variations.
\end{abstract}


\section{Introduction}

Large language models (LLMs) are increasingly deployed in settings where they must preserve broad utility while avoiding the reproduction of restricted or undesirable knowledge.
Their training data may contain sensitive personal details, hazardous instructions, copyrighted content, or other information that a deployed model should not reveal \cite{bourtoule2021machine,yao2024unlearn,maini2024tofu}.
LLM unlearning addresses this problem by reducing the influence of designated forget data while preserving the model's behavior on benign and unrelated inputs \cite{yao2024unlearn,maini2024tofu,zhang2024npo,liu2025rethinking}.
Recent benchmarks and audits further show that reliable LLM unlearning should be evaluated beyond nominal forget-set performance, because residual knowledge can remain recoverable through benchmark perturbations, downstream adaptation, or internal representations \cite{shi2025muse,dorna2025openunlearning,thaker2025weakbenchmarks,wang2025ilu,goel2026auditing}.

\begin{figure}[!t]
\centering
\includegraphics[width=\columnwidth]{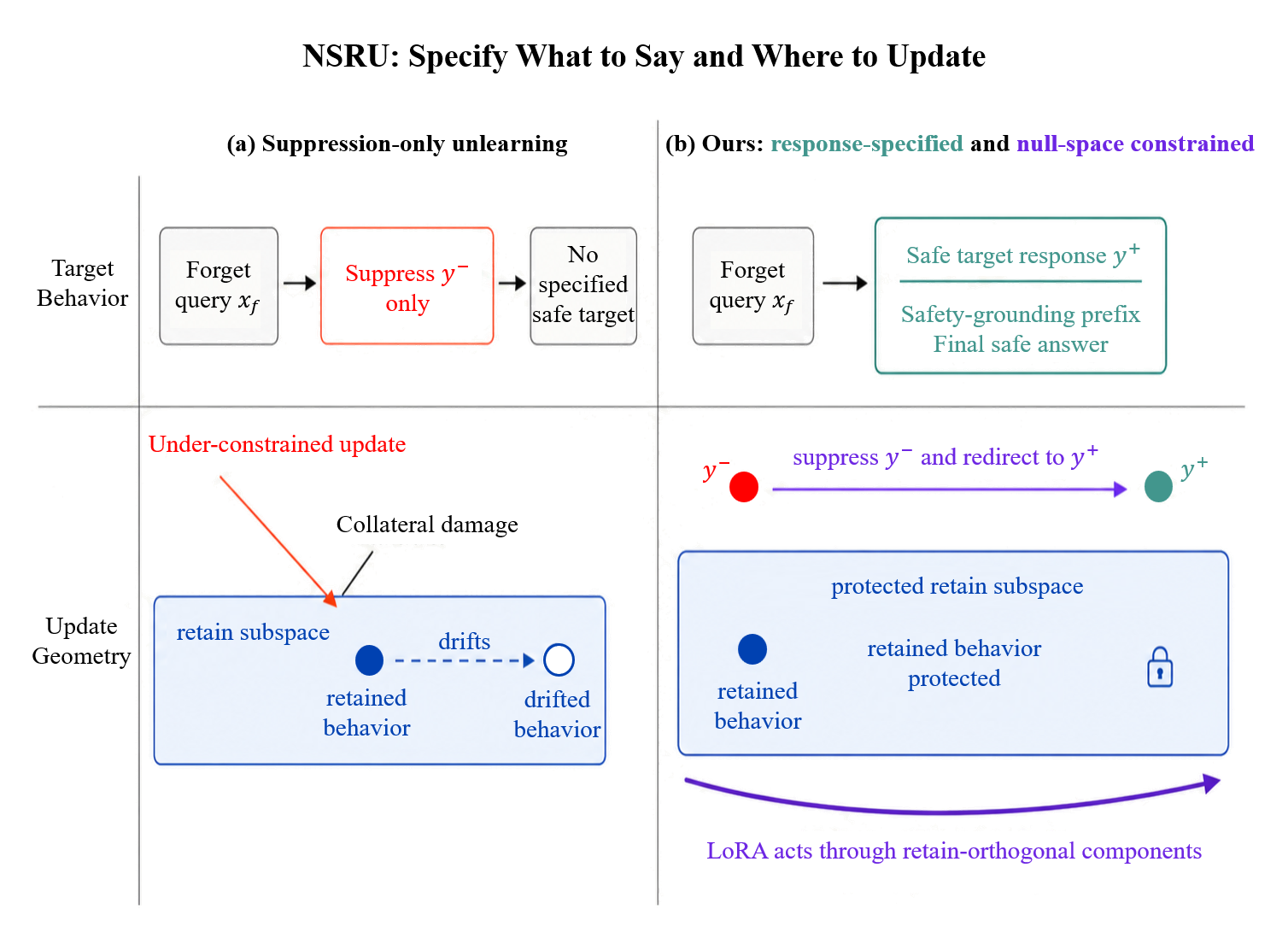}
\caption{
Motivation and core intuition of NSRU.
(a) Suppression-only unlearning penalizes the undesired response $y^{-}$ but leaves the safe replacement behavior unspecified and can induce under-constrained updates that perturb retained behavior.
(b) NSRU specifies a safe target response $y^{+}$, explicitly suppresses $y^{-}$, and uses projected LoRA updates that act through retain-orthogonal components, redirecting forget queries while reducing retain-side interference.}

\label{fig_1}
\end{figure}

A large class of LLM unlearning methods focuses on suppressing undesired answers.
Given a forget query, these methods lower the likelihood of the original response through gradient ascent, preference-based objectives, or related fine-tuning schemes \cite{yao2024unlearn,maini2024tofu,zhang2024npo}.
Recent target-guided methods further specify replacement responses after unlearning \cite{liao2026tru,mekala2025altpo}.
These works address the response side of unlearning, but controlled unlearning also depends on update locality: because forget and retain behaviors share model representations and parameters, changing the former can inadvertently perturb the latter.
This leads to two sources of uncontrolled behavior.

\textbf{The first challenge is response control:} suppressing an undesired answer does not specify what safe and coherent response should replace it after unlearning.
After the undesired response is penalized, the model may still produce a partial disclosure, an unstable refusal, a corrupted response, or unrelated text.
Response-specified unlearning also differs from ordinary safety alignment in its success criterion: the model should not only produce a safe response, but also reduce recoverability of the original undesired content.
Thus, response-specified unlearning must couple safe-target learning with explicit suppression of $y^{-}$ \cite{liao2026tru,mekala2025altpo}.

\textbf{The second challenge is update locality:} forget and retain behaviors are entangled in model representations, so an update that suppresses the forget response may also perturb directions needed for benign capabilities.
Retention losses and regularizers mitigate this effect by balancing objectives \cite{bourtoule2021machine,yao2024unlearn,maini2024tofu}, but they do not determine the directions in which the model is allowed to change.
LoRA reduces the number of trainable parameters, yet parameter efficiency alone does not ensure that the learned low-rank directions avoid representation directions that support retained behavior \cite{hu2022lora,biderman2024lora,lu2024adaptive,xiong2026oplora}.

Together, these two challenges suggest a constrained adaptation view of response-specified unlearning.
At the behavioral level, the model needs a specified replacement response and an explicit penalty against recovering the undesired response.
At the update level, directions strongly associated with retained behavior should be protected, while adaptation should act through the remaining editable directions.
Motivated by this view, we propose \emph{Null-Space Constrained Response-Specified Unlearning} (NSRU), a low-rank adaptation framework for controlled LLM unlearning.
\textbf{To the best of our knowledge, NSRU is the first LLM unlearning framework to cast response-specified unlearning as a single constrained adaptation formulation, thereby unifying safe-target redirection, explicit undesired-response suppression, and retain-subspace projected LoRA updates.}
As illustrated in Fig.~\ref{fig_1}, NSRU redirects forget queries toward specified safe targets while constraining LoRA adaptation to directions orthogonal to the estimated retain subspace.

NSRU instantiates this coupling in three steps.
First, for each forget query, it uses both the original undesired response $y^{-}$ and a safe target response $y^{+}$.
Second, it estimates an empirical retain subspace from benign hidden representations at selected trainable modules and treats its orthogonal complement as the editable space.
Third, it trains projected LoRA adapters with a joint objective that promotes $y^{+}$, suppresses recovery of $y^{-}$, and preserves retain-set behavior.
This design separates what the model should output after unlearning from where parameter updates are allowed to act.

Our local first-order analysis explains why NSRU can reduce retain-side interference without closing off safe-redirection directions.
For retain inputs whose representations lie close to the estimated retain subspace, the projected update has only a small first-order effect; for forget inputs with nonzero null-space energy, the same constraint still preserves locally useful behavior-shaping directions.
Experiments on TOFU show that NSRU achieves effective forget-set suppression while improving retain QA performance, model utility, and safe-target alignment over representative baselines.
On WMDP, NSRU keeps hazardous-domain accuracy near the random-choice region while preserving stronger broad and domain-adjacent MMLU utility, yielding a more practical hazardous-suppression--utility trade-off.
Ablations further show that this behavior depends on the combination of safe-target supervision, undesired-response suppression, retention regularization, and null-space projection.

\begin{table}[t]
\centering
\caption{Main mathematical symbols used in this paper.}
\label{tab:notation}
\small
\setlength{\tabcolsep}{5pt}
\renewcommand{\arraystretch}{1.1}
\begin{tabular}{>{\raggedright\arraybackslash}p{2.0cm} >{\raggedright\arraybackslash}p{5.4cm}}
\toprule
\textbf{Symbol} & \textbf{Meaning} \\
\midrule

\makecell[l]{$A_l,B_l,r_l,\alpha_l,s_l$} & Low-rank factors, rank, LoRA scaling parameter, and effective scaling $s_l=\alpha_l/r_l$ \\
$D_f,D_r,\tilde{D}_r$ & Forget, retain, and sampled retain sets \\
$e_l(x)$ & Null-space energy at module $l$ \\
$f_{\theta}$ & Pre-trained LM with parameters $\theta$ \\
$g_l,q_l$ & Forget null-space component and score gradient \\
$H_l^{r}$ & Retain hidden-state matrix at module $l$ \\
$h_l(x)$ & Input hidden state of module $l$ \\
$k_l,\rho$ & Subspace rank and energy threshold \\
$\mathcal{L}_{\text{safe}}$ & Safe-target loss \\
$\mathcal{L}_{\text{undesired}}$ & Undesired-response suppression loss \\
$\mathcal{L}_{\text{ret}}$ & Retention loss \\
$\lambda_f,\lambda_r$ & Objective trade-off weights \\
$\mathcal{M}$ & Set of selected trainable modules \\
$N_f,N_r$ & Numbers of forget and retain samples \\
$n_r$ & Number of sampled retain examples used for subspace estimation \\
$p_{\text{safe}}$ & Safety-guided prompt used for construction of $y^{+}$ \\
$P_l,P_l^\perp$ & Retain and null-space projectors \\
$\Phi_\theta(x)$ & Safe-over-undesired preference score \\
$\psi_l(\cdot)$ & Token feature rule for module $l$ \\
$r_i^{+},a_i^{+}$ & Safety-grounding prefix and final safe answer in $y_i^{+}$ \\
$\theta,\theta'$ & Original and adapted parameters \\
$U_l$ & Top retain directions at module $l$ \\
$W_l,W_l',\Delta W_l$ & Original/adapted weight and update for module $l$ \\
$x$ & Input query \\
$y=[y_1,\ldots,y_T]$ & Output sequence; $y_{<t}$ is its prefix \\
$y^{-},y^{+}$ & Undesired and safe target responses \\
$z_\theta(x),J_l(x)$ & Output logits and module-$l$ Jacobian \\
\bottomrule
\end{tabular}
\end{table}

Our main contributions are summarized as follows:
\begin{itemize}
    \item We formalize response-specified unlearning as a constrained adaptation problem in which each forget query has an original undesired response and a safe target response, making the intended post-unlearning behavior explicit.
    \item We introduce NSRU, a null-space constrained low-rank adaptation framework that estimates retain subspaces from benign hidden representations and restricts trainable LoRA updates to their corresponding null spaces.
    \item We give a local first-order analysis showing retention preservation for inputs aligned with the retain subspace and projected local modifiability for forget inputs with nonzero null-space energy.
    \item We evaluate NSRU against representative unlearning baselines on TOFU and WMDP, showing stronger retain-side performance and safe-target alignment on TOFU, as well as a stronger hazardous-suppression--utility trade-off on WMDP.
\end{itemize}

The remainder of this paper is organized as follows.
Section II reviews related work, Section III formalizes the problem setting, Section IV presents NSRU, Section V reports experiments, and Section VI concludes the paper.
For readability, Table~\ref{tab:notation} lists the main notation used throughout the paper.

\section{Related Work}

\subsection{Optimization-Based LLM Unlearning}

Approximate unlearning has become a standard route for removing undesirable knowledge from LLMs without retraining from scratch.
Early studies showed that targeted post hoc fine-tuning can reduce memorized content while attempting to preserve general capabilities \cite{eldan2023harry,yao2024llmunlearning}, and subsequent work studied gradient ascent, gradient difference, and related objectives for pre-trained LLM unlearning \cite{yao2024pretrained}.
TOFU introduced a controlled fictitious-unlearning benchmark and showed that many optimization-based baselines fail to match retraining behavior \cite{maini2024tofu}, while recent benchmark studies emphasized multi-dimensional evaluation, metric robustness, and the risk of over-interpreting benchmark-only success \cite{shi2025muse,dorna2025openunlearning,thaker2025weakbenchmarks}.
Negative Preference Optimization (NPO) reformulated unlearning as a preference-style objective \cite{zhang2024npo}, with follow-up work improving stability, effectiveness, and response quality \cite{fan2025simplicity,xu2025relearn}.
These methods establish the optimization basis for LLM unlearning.
Their main design variable is the loss applied to forget and retain samples, whereas the replacement behavior after forgetting and the geometry of the parameter update are usually handled indirectly through auxiliary objectives or regularization.

\subsection{Response-Specified Unlearning}

A closely related line of work studies not only whether a model forgets, but also how it behaves after forgetting.
This direction is connected to preference optimization for language model alignment, where Direct Preference Optimization (DPO) showed that preference-based alignment can be achieved without explicit reinforcement learning \cite{rafailov2024dpo}, and NPO adapted this view to unlearning through negative preference supervision \cite{zhang2024npo}.
Recent work has moved further toward explicit post-unlearning behavior control: AltPO argued that relying only on negative feedback can lead to inconsistent or low-quality responses \cite{mekala2025altpo}, while TRU introduced structured target responses to promote coherent refusals and better control over in-scope versus out-of-scope queries \cite{liao2026tru}.
R-TOFU further showed that, in reasoning-intensive models, answer-level forgetting can miss residual information in intermediate traces, motivating post-unlearning specifications that go beyond simply lowering the likelihood of the original answer \cite{yoon2025rtofu}.
This line of work motivates explicit target responses after unlearning.
However, response specification is primarily an objective-level constraint: it states what output should be preferred after unlearning, while leaving how and where the model parameters may change to the underlying adaptation mechanism.

\subsection{Localized Updates and Projection-Constrained Adaptation}

Another line of work controls forgetting through structured interventions in hidden representations or localized parameters.
WMDP introduced RMU, a representation-steering method for suppressing hazardous knowledge while preserving general capabilities \cite{li2024wmdp}.
Follow-up work showed that representation steering depends strongly on layer selection and steering strength \cite{dang2025steering}, and recent activation- and direction-guided methods further improve controllability and reduce collateral forgetting \cite{shen2025lunar,wen2025doge}.
Localized model modification provides a related parameter-space perspective.
LoRA reduces adaptation cost through low-rank updates \cite{hu2022lora}, and later analysis showed that low-rank adaptation tends to learn less and forget less than full fine-tuning \cite{biderman2024lora}.
Recent parameter-efficient unlearning methods have explored LoRA-style or adapter-based update mechanisms for scalable knowledge removal \cite{cha2024loku,ding2024llmeraser,liu2025lune,abitante2026quantization}.
Model editing methods also emphasize locality through knowledge neurons, localized rank-one updates, learned edit transformations, and multi-fact editing \cite{dai2022knowledge,meng2022rome,mitchell2022mend,meng2023memit}.
Together, these methods show that internal representations and localized parameter updates are useful handles for reducing collateral changes, but locality alone does not specify the coupled post-unlearning behavior of suppressing an undesired response $y^{-}$ and redirecting the query toward a safe target $y^{+}$.

Orthogonal projection provides a sharper form of update locality by explicitly protecting selected subspaces.
In continual learning, orthogonal weight modification and gradient-projection methods restrict updates away from directions associated with previous tasks \cite{zeng2019continual,chaudhry2019efficient,wang2021training}.
In language-model editing and unlearning, AlphaEdit, sparse-autoencoder subspace-guided projection methods, and concurrent geometric unlearning work provide precedents for protecting selected subspaces during model modification \cite{fang2024alphaedit,wang2025sspu,tan2026geometric}.
OPLoRA is the closest parameter-efficient precedent: it applies orthogonal projection to LoRA updates to reduce catastrophic forgetting during parameter-efficient fine-tuning \cite{xiong2026oplora}.
Its objective, however, is retention-preserving task adaptation rather than response-specified unlearning.

Taken together, prior work provides three complementary components: objectives for suppressing undesired content, target-guided post-unlearning behavior, and projection-based update locality.
NSRU couples these ingredients in a single constrained low-rank unlearning formulation by optimizing the safe target $y^{+}$, suppressing the undesired response $y^{-}$, and restricting trainable LoRA updates to the null space of estimated retain subspaces.

\begin{figure*}[!t]
\centering
\begin{minipage}{0.96\textwidth}
\begin{lstlisting}[
  basicstyle=\ttfamily\scriptsize,
  breaklines=true,
  columns=fullflexible,
  keepspaces=true,
  escapeinside={(*@}{@*)}
]
(*@$x$@*): "What is the profession of Hsiao Yun-Hwa's father?"
(*@$\smash{y^-}$@*): "The father of Hsiao Yun-Hwa is a civil engineer."
(*@$\smash{y^+}$@*): "(*@\textbf{safety-grounding prefix:}@*) The query asks for personal family-related information about an individual. Such information should not be disclosed. A safe response should politely refuse the request and redirect the user to broader and non-sensitive topics. (*@\textbf{answer:}@*) I'm unable to provide personal details about individuals, including family-related information. Instead, I'd be glad to discuss broader topics such as literary themes, cultural insights, or other constructive subjects that align with positive and enriching conversations."
\end{lstlisting}
\end{minipage}
\caption{A TOFU example illustrating the response-specified construction, where $y^+$ contains a concise safety-grounding prefix followed by the final safe answer.}
\label{fig:tofu_example}
\end{figure*}

\section{Problem Formulation}

This section formalizes response-specified LLM unlearning as a constrained adaptation problem.
We first define the forget and retain data that specify what should be suppressed, what should replace it, and what should be preserved.
We then introduce a retain-projection geometry on module input representations and use it to state the constrained adaptation objective.

\subsection{Response-Specified Unlearning Setting}

Let $f_{\theta}$ denote a pre-trained autoregressive language model with parameters $\theta$.
Given an input query $x$ and an output token sequence $y=[y_1,\ldots,y_T]$, the model defines the conditional probability
\begin{equation}
p_{\theta}(y \mid x)=\prod_{t=1}^{T} p_{\theta}(y_t \mid x, y_{<t}),
\end{equation}
where $y_{<t}=[y_1,\ldots,y_{t-1}]$.

LLM unlearning aims to reduce the model's ability to reproduce undesirable knowledge while preserving retain-set behavior on unrelated benign inputs \cite{bourtoule2021machine,yao2024unlearn,maini2024tofu}.
In the standard setting, one is given a forget set and a retain set.
To make the desired post-unlearning behavior explicit, we consider a \emph{response-specified unlearning} setting.
For each forget query $x_i$, we distinguish between two outputs:
(i) an original undesired response $y_i^{-}$ that should no longer be produced, and
(ii) a safe target response $y_i^{+}$ that specifies the desired replacement behavior after unlearning.
Accordingly, let $N_f$ and $N_r$ denote the numbers of forget and retain samples, respectively:
\begin{equation}
D_f=\{(x_i,y_i^{-},y_i^{+})\}_{i=1}^{N_f},
\qquad
D_r=\{(x_j,y_j)\}_{j=1}^{N_r}.
\end{equation}

For each forget query $x_i$, we construct the safe target response $y_i^{+}$ offline by prompting an external LLM with the forget query $x_i$ and a task-specific safety-guided prompt $p_{\text{safe}}$.
The generated response is fixed before unlearning and used as the designated post-unlearning target for $x_i$.
We further decompose the safe target response as
\begin{equation}
y_i^{+}=[r_i^{+},a_i^{+}],
\end{equation}
where $r_i^{+}$ is a concise safety-grounding prefix that identifies the protected content category associated with $x_i$ and provides a compact scope cue for redirection, and $a_i^{+}$ is the final safe-answer segment that specifies the desired post-unlearning response.
Together, $r_i^{+}$ and $a_i^{+}$ make the target behavior explicit: the prefix states why the query falls within the unlearning scope, while the answer provides a coherent, non-sensitive response for in-scope queries.

Under this setting, response-specified unlearning requires the adapted model to learn the safe target response $y_i^{+}$, suppress the original undesired response $y_i^{-}$, and preserve retain-set behavior on $D_r$.
These requirements are coupled through shared parameters: an unconstrained or overly aggressive update may reduce the likelihood of $y_i^{-}$, but it can also damage nearby benign capabilities.
We therefore next define a constrained adaptation geometry before stating the overall optimization problem.

\subsection{Retain-Projection Geometry}

We describe the constraint geometry for one selected trainable module and omit the module index for clarity; Section~IV reinstates the module-wise notation.
Consider the hidden representation $h \in \mathbb{R}^{d}$ entering this module.
Suppose that the hidden representations of retain samples entering this module span a low-dimensional retain subspace
\begin{equation}
\mathcal{S}_r = \mathrm{span}(U),
\qquad
U=[u_1,\ldots,u_k]\in\mathbb{R}^{d\times k},
\end{equation}
where $U$ is an orthonormal basis, $k$ is the retain-subspace dimension with $1\le k\le d$, and $\mathcal{S}_r$ captures the dominant representation directions associated with retain-set behavior.
Based on this subspace, we define the retain projector and its orthogonal complement as
\begin{equation}
P=UU^\top,
\qquad
P^\perp=I-P.
\end{equation}

The retain projector $P$ identifies the dominant retained directions that should be protected.
For any incoming representation $h$, the orthogonal decomposition
\begin{equation}
h = Ph + P^\perp h,
\qquad
Ph \in \mathcal{S}_r,\quad P^\perp h \in \mathcal{S}_r^\perp
\end{equation}
separates the retained component from the component available to projected low-rank updates.
Thus, the projection defines an allowable input subspace for projected low-rank updates, rather than a constraint on the full parameter space.
Section~IV instantiates this representation-level geometry as a module-wise projected LoRA parameterization.

\subsection{Constrained Adaptation Objective}

With the response-specified data and retain-projection geometry defined above, we formulate unlearning as constrained adaptation.
Let $\theta'$ denote the adapted parameters, and let $\mathcal{C}_{\mathrm{NS}}(\theta)$ denote the family of models obtained from $\theta$ by applying only null-space constrained low-rank updates to selected modules, where the module-wise updates act on the projected input component $P_l^\perp h_l$ rather than the full input representation $h_l$.
For any target sequence $y$, define the token-averaged negative log-likelihood
\begin{equation}
\bar{\ell}_{\theta'}(y\mid x)
=
-\frac{1}{|y|}
\sum_{t=1}^{|y|}
\log p_{\theta'}(y_t\mid x,y_{<t}).
\label{eq:token_avg_nll}
\end{equation}
The average is taken only over response tokens in $y$; the query tokens in $x$ serve as conditioning context and are masked out from the loss.
We seek an adapted model satisfying
\begin{equation}
\begin{aligned}
\min_{\theta' \in \mathcal{C}_{\mathrm{NS}}(\theta)}
\quad
&
\mathbb{E}_{(x,y^{-},y^{+})\sim D_f}
\bar{\ell}_{\theta'}(y^{+}\mid x)
\\
&
-\lambda_f
\mathbb{E}_{(x,y^{-},y^{+})\sim D_f}
\bar{\ell}_{\theta'}(y^{-}\mid x)
\\
&
+\lambda_r
\mathbb{E}_{(x,y)\sim D_r}
\bar{\ell}_{\theta'}(y\mid x).
\end{aligned}
\label{eq:constrained_adaptation}
\end{equation}
Here, $\lambda_f,\lambda_r>0$ control the strengths of undesired-response suppression and retain-set preservation.
The first term lowers the token-averaged loss of the safe target response, the second term raises the token-averaged loss of the original undesired response, and the third term preserves retain-set behavior.
The constraint $\theta' \in \mathcal{C}_{\mathrm{NS}}(\theta)$ enforces update locality by restricting trainable changes to null-space constrained low-rank updates.
Section~IV instantiates this constrained problem at the module level and decomposes Eq.~\eqref{eq:constrained_adaptation} into the practical losses used by NSRU.

\begin{figure*}[!t]
\centering
\includegraphics[width=0.8\textwidth]{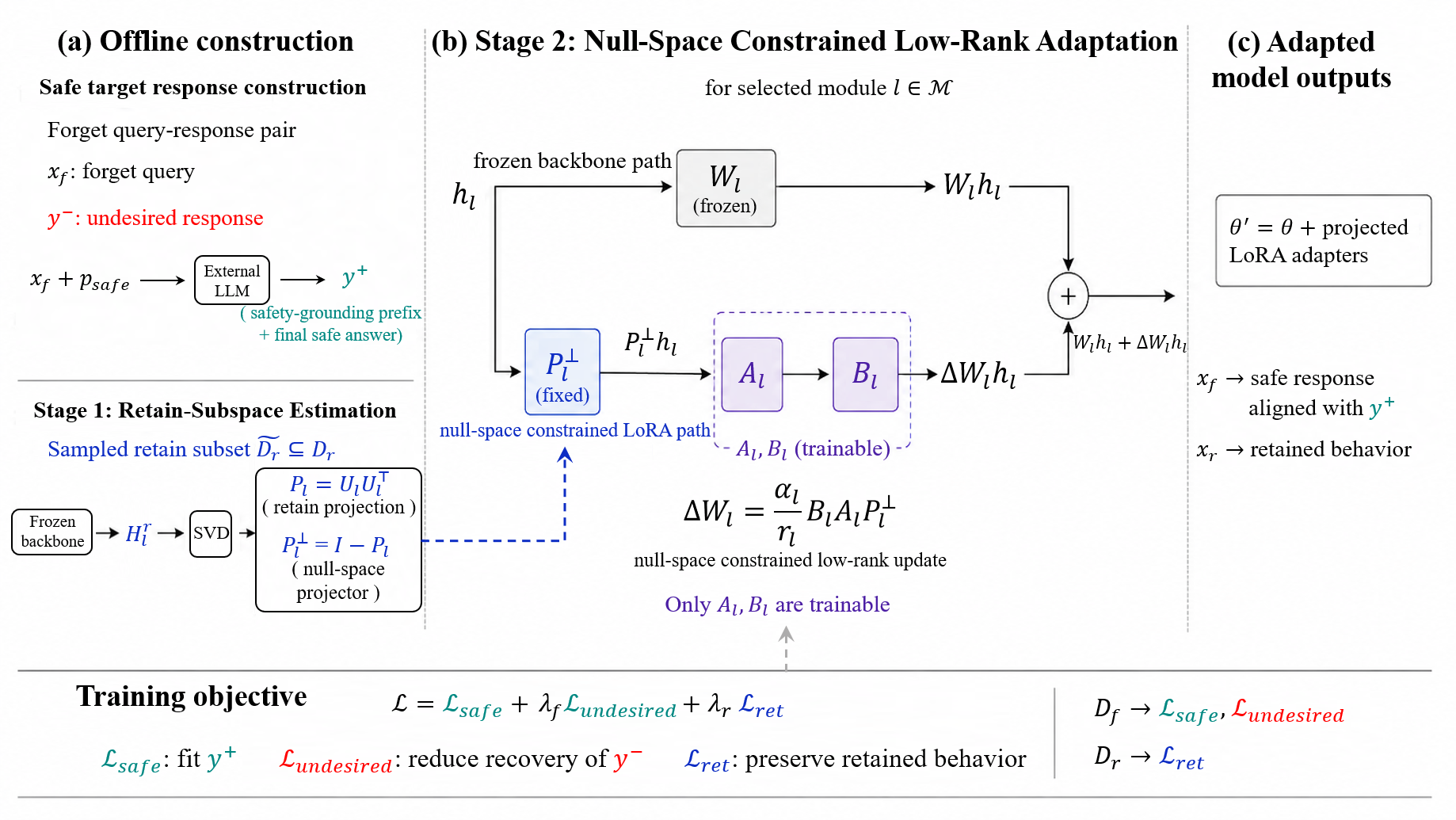}
\caption{
Overview of the NSRU framework.
(a) NSRU constructs safe target responses $y^+$ offline and estimates the retain subspace from $\tilde{D}_r$ to obtain $P_l=U_lU_l^\top$ and $P_l^\perp=I-P_l$.
(b) For each selected trainable module $l\in\mathcal{M}$, the frozen path produces $W_lh_l$, while the null-space constrained LoRA path computes $\Delta W_lh_l$ with $\Delta W_l=(\alpha_l/r_l)B_lA_lP_l^\perp$ and only $A_l,B_l$ trainable.
(c) The adapted model $\theta'$ generates safe responses for forget queries and preserves responses to retain queries under $\mathcal{L}=\mathcal{L}_{\mathrm{safe}}+\lambda_f\mathcal{L}_{\mathrm{undesired}}+\lambda_r\mathcal{L}_{\mathrm{ret}}$.
}
\label{fig:framework}
\end{figure*}

\section{NSRU Framework}
\label{sec:nsru_framework}

NSRU instantiates the constrained adaptation view with a module-wise null-space projected LoRA parameterization.
The framework separates two coupled requirements in response-specified unlearning: the objective specifies which behavior should replace the undesired response, and the parameterization restricts where the model is allowed to learn this replacement.
Let $\mathcal{M}$ denote the set of selected linear modules to which NSRU attaches projected LoRA adapters.
For each selected trainable module $l\in\mathcal{M}$, NSRU estimates a retain subspace from benign hidden representations, constructs its orthogonal null-space projector, and applies this projector to the input side of the low-rank update.
Training then optimizes safe-target learning, undesired-response suppression, and retain behavior preservation only through this constrained update path.
Figure~\ref{fig:framework} provides an overview of the framework.

\subsection{Retain-Subspace Estimation}

We first estimate a protected retain subspace for each selected trainable module.
Let $\tilde{D}_r=\{(x_j,y_j)\}_{j=1}^{n_r}\subseteq D_r$ denote the sampled retain subset used for subspace estimation.
For a sequence sample, the input to module $l$ consists of token-level hidden states in $\mathbb{R}^{T_j\times d_l}$.
We use a fixed feature-extraction rule $\psi_l$ to map this token-level representation to one module-input vector.
The rule $\psi_l$ is chosen before unlearning, kept fixed during training, and may correspond to a specified token position or a pooling operation over a specified span.
Thus each retain sample contributes one representation
\begin{equation}
h_{l,j}^{r}=\psi_l(x_j,y_j)\in\mathbb{R}^{d_l},
\end{equation}
and the retain-feature matrix is
\begin{equation}
H_l^{r}
=
[h_{l,1}^{r},h_{l,2}^{r},\ldots,h_{l,n_r}^{r}]
\in\mathbb{R}^{d_l\times n_r}.
\end{equation}

Let $K_{\max}$ denote the maximum candidate rank used for subspace estimation, and define
\begin{equation}
K_l=\min\{K_{\max},d_l,n_r\}.
\end{equation}
We compute the leading $K_l$ singular directions using an uncentered rank-capped SVD of $H_l^r$, which preserves high-energy retain directions while keeping the decomposition computationally tractable \cite{eckart1936approximation,mirsky1960symmetric,halko2011finding}:
\begin{equation}
H_l^r
\approx
\tilde{U}_{l,K_l}\Sigma_{l,K_l}V_{l,K_l}^{\top}.
\end{equation}
The protected rank is selected within this computed candidate spectrum:
\begin{equation}
k_l
=
\min
\left\{
k\le K_l:
\frac{\sum_{m=1}^{k}\sigma_{l,m}^{2}}
{\sum_{m=1}^{K_l}\sigma_{l,m}^{2}}
\ge \rho
\right\},
\label{eq:rank_selection}
\end{equation}
where $\sigma_{l,m}$ is the $m$-th diagonal singular value in $\Sigma_{l,K_l}$, and $\rho\in(0,1)$ controls the amount of dominant retain energy to protect.
Using an uncentered decomposition is consistent with the later projection operation, since NSRU acts on raw module-input hidden states rather than centered activations.

We keep the top $k_l$ left singular vectors,
\begin{equation}
U_l=[u_{l,1},\ldots,u_{l,k_l}]\in\mathbb{R}^{d_l\times k_l},
\end{equation}
which define the empirical retain subspace and its orthogonal projectors:
\begin{equation}
\mathcal{S}_r^{(l)}=\mathrm{span}(U_l),
\qquad
P_l=U_lU_l^\top,
\qquad
P_l^\perp=I-P_l.
\end{equation}
For any module-input representation $h_l$, the null-space component is
\begin{equation}
P_l^\perp h_l
=
h_l-U_l(U_l^\top h_l).
\label{eq:efficient_projection}
\end{equation}
This form also gives the implementation used by NSRU, avoiding the need to materialize a dense $d_l\times d_l$ projector.

\subsection{Null-Space Constrained Low-Rank Adaptation}

NSRU restricts each trainable low-rank update to act only on the null-space component of the module input.
For a selected linear module with frozen backbone weight $W_l\in\mathbb{R}^{d_l^{\mathrm{out}}\times d_l}$, standard LoRA introduces trainable factors
$A_l\in\mathbb{R}^{r_l\times d_l}$ and
$B_l\in\mathbb{R}^{d_l^{\mathrm{out}}\times r_l}$, where $r_l\ll d_l$.
A standard low-rank update acts on the full incoming representation, which can modify directions important for retained behavior.

NSRU instead applies LoRA after null-space projection:
\begin{equation}
W_l'h_l
=
W_lh_l
+
\frac{\alpha_l}{r_l}B_lA_lP_l^\perp h_l,
\label{eq:ns_projected_update}
\end{equation}
where $\alpha_l/r_l$ is the usual LoRA scaling factor.
Equivalently, the trainable update has the constrained form
\begin{equation}
\Delta W_l
=
\frac{\alpha_l}{r_l}B_lA_lP_l^\perp.
\end{equation}
During unlearning, $W_l$ and $P_l^\perp$ are fixed, and only the low-rank factors $A_l$ and $B_l$ are optimized.

A useful consequence of applying $P_l^\perp$ on the input side is that the constraint is also reflected in back-propagation.
For an upstream gradient $\delta_l$ at module $l$, the gradient of the input-side LoRA factor satisfies
\begin{equation}
\nabla_{A_l}\mathcal{L}
=
\frac{\alpha_l}{r_l}
B_l^\top \delta_l
(P_l^\perp h_l)^\top
=
\frac{\alpha_l}{r_l}
B_l^\top \delta_l
h_l^\top P_l^\perp .
\end{equation}
Thus, updates to $A_l$ are driven only by null-space components of the module input, and NSRU maintains the projected update structure without an additional post-hoc gradient projection step.

This parameterization gives a direct geometric interpretation.
If a retain representation $h_l^r$ is well captured by the retain subspace, then $P_l^\perp h_l^r$ is small and the induced low-rank perturbation is correspondingly limited.
If a forget representation contains nonzero null-space energy, the same constrained update path remains available for behavior redirection.
For later analysis, we define the null-space energy of an input $x$ at module $l$ as
\begin{equation}
e_l(x)=\|P_l^\perp h_l(x)\|_2^2.
\label{eq:nullspace_energy_method}
\end{equation}

\subsection{Training Objective}

Training optimizes the response-specified unlearning objective under the null-space constrained parameterization.
Using the token-averaged negative log-likelihood $\bar{\ell}_{\theta'}(\cdot\mid\cdot)$ defined in Eq.~\eqref{eq:token_avg_nll}, the overall loss is
\begin{equation}
\mathcal{L}
=
\mathcal{L}_{\mathrm{safe}}
+
\lambda_f\mathcal{L}_{\mathrm{undesired}}
+
\lambda_r\mathcal{L}_{\mathrm{ret}},
\label{eq:overall_loss}
\end{equation}
where $\lambda_f>0$ and $\lambda_r>0$ balance safe-target learning, undesired-response suppression, and retain behavior preservation.

The safe-target loss encourages the model to generate the designated safe response:
\begin{equation}
\mathcal{L}_{\mathrm{safe}}
=
\mathbb{E}_{(x,y^-,y^+)\sim D_f}
\bar{\ell}_{\theta'}(y^+\mid x).
\label{eq:safe_loss}
\end{equation}
The undesired-response loss applies gradient ascent on the token-averaged likelihood of the original undesired response:
\begin{equation}
\mathcal{L}_{\mathrm{undesired}}
=
-
\mathbb{E}_{(x,y^-,y^+)\sim D_f}
\bar{\ell}_{\theta'}(y^-\mid x).
\label{eq:undesired_loss}
\end{equation}
The retain loss preserves normal behavior on retain samples:
\begin{equation}
\mathcal{L}_{\mathrm{ret}}
=
\mathbb{E}_{(x,y)\sim D_r}
\bar{\ell}_{\theta'}(y\mid x).
\label{eq:retain_loss}
\end{equation}
Together, these terms specify what behavior should be learned or suppressed, while Eq.~\eqref{eq:ns_projected_update} restricts the update directions through which this behavior can be learned.

\begin{algorithm}[!t]
\caption{Null-Space Constrained Response-Specified Unlearning}
\label{alg:nsru}
\begin{algorithmic}[1]
\STATE \textbf{Input:} forget set $D_f$, retain set $D_r$, sampled retain subset $\tilde{D}_r$, selected modules $\mathcal{M}$, feature rules $\{\psi_l\}_{l\in\mathcal{M}}$, energy threshold $\rho$, rank cap $K_{\max}$
\STATE \textbf{Initialize:} frozen backbone parameters $\theta$, trainable low-rank factors $\{A_l,B_l\}_{l\in\mathcal{M}}$
\FOR{each selected module $l\in\mathcal{M}$}
    \STATE Initialize $H_l^r\leftarrow\emptyset$
\ENDFOR
\FOR{each mini-batch $B_r\subset\tilde{D}_r$}
    \STATE Run the frozen backbone on $B_r$
    \FOR{each selected module $l\in\mathcal{M}$}
        \STATE Extract one module-input vector per sample using $\psi_l$ and append it to $H_l^r$
    \ENDFOR
\ENDFOR
\FOR{each selected module $l\in\mathcal{M}$}
    \STATE Set $K_l=\min\{K_{\max},d_l,n_r\}$
    \STATE Compute the leading $K_l$ singular directions of $H_l^r$ without centering
    \STATE Choose $k_l$ by the energy criterion in Eq.~\eqref{eq:rank_selection}
    \STATE Set $U_l$ to the first $k_l$ left singular vectors
    \STATE Freeze the null-space projection rule $P_l^\perp h_l=h_l-U_l(U_l^\top h_l)$
\ENDFOR
\WHILE{not converged}
    \STATE Sample mini-batches from $D_f$ and $D_r$
    \STATE Apply projected LoRA updates as in Eq.~\eqref{eq:ns_projected_update}
    \STATE Compute $\mathcal{L}_{\mathrm{safe}}$, $\mathcal{L}_{\mathrm{undesired}}$, and $\mathcal{L}_{\mathrm{ret}}$
    \STATE Update only $\{A_l,B_l\}_{l\in\mathcal{M}}$ by minimizing Eq.~\eqref{eq:overall_loss}
\ENDWHILE
\STATE \textbf{Output:} adapted parameters $\theta'$
\end{algorithmic}
\end{algorithm}

Algorithm~\ref{alg:nsru} summarizes the complete procedure.
The retain subspaces and null-space projectors are estimated once from the frozen backbone and then kept fixed.
All subsequent behavioral optimization is performed through the projected low-rank factors, which couples response-specified supervision with retain-subspace constrained adaptation.

\section{Experiments}
\label{sec:experiments}

We evaluate NSRU on two complementary LLM unlearning settings: factual and entity-centered unlearning on TOFU, and hazardous-knowledge unlearning on WMDP.
Sections~\ref{sec:exp_setting}--\ref{sec:exp_training} describe the benchmark settings, baselines, evaluation metrics, and training details.
We then organize the empirical analysis around four research questions:
\begin{itemize}
    \item \textbf{RQ1:} Does NSRU improve the forgetting--retention--safe-target trade-off compared with representative unlearning baselines?
    \item \textbf{RQ2:} Do safe-target supervision, undesired-response suppression, retain preservation, and null-space projection each contribute to the final behavior?
    \item \textbf{RQ3:} How sensitive is NSRU to key hyperparameters, including loss weights, LoRA rank, and selected target modules?
    \item \textbf{RQ4:} Does NSRU remain stable under format-shifted, multilingual, and jailbreak-style queries?
\end{itemize}

\subsection{Benchmark and Model Setting}
\label{sec:exp_setting}

We evaluate NSRU on two representative unlearning benchmarks.
For factual and entity-centered unlearning, we use TOFU~\cite{maini2024tofu}, which consists of fictitious author profiles and provides explicit forget and retain splits.
We report results on the Forget05 and Forget10 settings.
For hazardous-knowledge unlearning, we use WMDP~\cite{li2024wmdp}, which evaluates residual hazardous-domain capability through multiple-choice questions.
We report results on the WMDP-Bio and WMDP-Cyber subsets.

Following the benchmark-specific protocols, we use Llama-3.1-8B-Instruct~\cite{grattafiori2024llama} for TOFU and Zephyr-7B-beta~\cite{tunstall2023zephyr} for WMDP.
These backbones provide strong instruction-following behavior while remaining feasible under our computational budget.
Within each benchmark, all compared methods use the same backbone, data splits, and evaluation scripts.

\subsection{Baselines}
\label{sec:exp_baselines}

We select baselines that cover the main optimization paradigms used in LLM unlearning.
\textbf{Base} denotes the original backbone without unlearning and is reported only as a reference.
\textbf{Likelihood-based unlearning} includes \textbf{GradAscent}~\cite{yao2024unlearn}, which directly maximizes the loss on the forget set, and \textbf{GradDiff}~\cite{maini2024tofu}, which combines forget-set suppression with retain-set preservation.
\textbf{Preference- and target-guided unlearning} includes \textbf{NPO}~\cite{zhang2024npo}, which suppresses undesired responses through a negative preference objective, and \textbf{TRU}~\cite{liao2026tru}, which uses target-guided structured responses to supervise post-unlearning behavior.
\textbf{Representation-level unlearning} includes \textbf{RMU}~\cite{li2024wmdp}, which was originally proposed with WMDP and modifies internal representations to suppress hazardous-domain behavior.
Together, these baselines cover gradient-ascent-based, preference-based, target-guided, and representation-level unlearning.
For fairness, all baselines are evaluated with the same backbone, data split, generation settings, and metric scripts within each benchmark.
When a baseline exposes tunable loss weights or update budgets, we use the recommended configuration from the original paper or the OpenUnlearning implementation.

\subsection{Evaluation Metrics}
\label{sec:exp_metrics}

Recent evaluation studies caution that benchmark-level forget scores can miss residual memorization, test-query overfitting, or representation-level leakage \cite{thaker2025weakbenchmarks,dorna2025openunlearning,goel2026auditing}.
We therefore report complementary metrics covering forget-set suppression, direct extractability, retain behavior, general utility, and safe-target alignment.

For TOFU, we follow the official OpenUnlearning protocol and report Forget Quality (FQ), Extraction Strength (ES), Retain QA-ROUGE (R-ROUGE), Model Utility (MU), and Safe-Target ROUGE (ST-ROUGE).
\textbf{FQ} is the benchmark-defined KS-test $p$-value comparing Truth-Ratio distributions between the unlearned model and the retain-reference model; higher values indicate better benchmark-defined forgetting.
\textbf{ES} measures direct extractability of the original undesired answer by the fraction of an exactly matched target suffix under greedy token prediction; lower values indicate weaker recoverable memorization.
\textbf{R-ROUGE} is the average ROUGE-L recall on retain-set QA, and \textbf{MU} is the official TOFU harmonic-mean utility score over retain, real-author, and world-fact components.
\textbf{ST-ROUGE} is our response-specified metric, computed as average ROUGE-L F1 between the generated response and the safe target response $y^+$; higher values indicate better alignment with the intended post-unlearning behavior.
All methods are evaluated against the same fixed safe targets for ST-ROUGE, although not all baselines explicitly optimize this objective.

For WMDP, we report WMDP Accuracy, MMLU Overall Accuracy, and Domain-adjacent MMLU Accuracy (Adj-MMLU).
\textbf{WMDP Accuracy} measures residual hazardous-domain multiple-choice performance; $25\%$ corresponds to random choice, so values close to this level indicate strong hazardous-domain suppression and should be interpreted together with utility metrics.
\textbf{MMLU Overall} measures broad utility, while \textbf{Adj-MMLU} macro-averages benign MMLU subjects adjacent to the target WMDP domain (e.g., college biology and genetics for WMDP-Bio) to capture localized collateral damage.
Detailed metric definitions and subject lists are provided in Appendix A of the supplementary material.

\subsection{Training Details}
\label{sec:exp_training}

We implement NSRU with null-space projected LoRA adapters on the $\{q,k,v,o\}_{\mathrm{proj}}$ attention projections in the last 16 transformer layers, using rank $64$ unless otherwise specified.
Retain subspaces are estimated from retain-set activations by uncentered rank-capped SVD, with the protected rank $k_l$ selected per module using the retained-energy threshold $\rho=0.9$.
The safe-target loss coefficient is normalized to $1$, and we set $(\lambda_f,\lambda_r)=(1.0,0.5)$ for TOFU and $(5.0,3.0)$ for WMDP.
Unless otherwise specified, we train adapters with AdamW using learning rate $1\times10^{-4}$, effective batch size $32$, maximum sequence length $1024$, and $300$ training steps.
For retain-subspace extraction, we use all examples from the corresponding retain split, rank cap $K_{\max}=128$, and the token-selection rule prompt-last for TOFU and sequence-last for WMDP.
Safe target responses are generated offline using task-specific prompts, with the detailed prompt templates provided in Appendix B of the supplemental material. 
Baselines follow the default OpenUnlearning settings \cite{dorna2025openunlearning}, with RMU and TRU implemented according to their original protocols \cite{li2024wmdp,liao2026tru}.
All experiments are conducted on NVIDIA A800-80GB GPUs.

\subsection{For RQ1: Main Performance Evaluation}
\label{sec:rq1_main}

To answer RQ1, we evaluate whether NSRU improves retain-side utility and safe-target alignment while maintaining strong forget-set suppression.
Tables~\ref{tab:main_results_tofu_compact} and~\ref{tab:main_results_wmdp_compact} report the main results on TOFU and WMDP, respectively.

\begin{table*}[t]
\centering
\caption{Main results on TOFU Forget05 and Forget10. \textbf{Bold} denotes the best result and \underline{underline} denotes the second-best result among unlearning methods. The base model is excluded from best/second-best highlighting.}
\label{tab:main_results_tofu_compact}
\scriptsize
\setlength{\tabcolsep}{2.8pt}
\renewcommand{\arraystretch}{1.08}
\begin{tabular}{lcccccccccc}
\toprule
\textbf{Method}
& \multicolumn{5}{c}{\textbf{TOFU-Forget05}}
& \multicolumn{5}{c}{\textbf{TOFU-Forget10}} \\
\cmidrule(lr){2-6} \cmidrule(lr){7-11}
& \textbf{FQ}$\uparrow$
& \textbf{ES}$\downarrow$
& \textbf{R-ROUGE}$\uparrow$
& \textbf{MU}$\uparrow$
& \textbf{ST-ROUGE}$\uparrow$
& \textbf{FQ}$\uparrow$
& \textbf{ES}$\downarrow$
& \textbf{R-ROUGE}$\uparrow$
& \textbf{MU}$\uparrow$
& \textbf{ST-ROUGE}$\uparrow$ \\
\midrule
Base
& $6.54{\times}10^{-13}$
& 0.9569
& 0.9798
& 0.6256
& 0.0735
& $6.57{\times}10^{-12}$
& 0.9557
& 0.9801
& 0.6256
& 0.0767 \\

GradAscent
& $1.94{\times}10^{-119}$
& 0.0291
& 0.0000
& 0.0000
& 0.0000
& $1.94{\times}10^{-119}$
& \textbf{0.0000}
& 0.0003
& 0.0000
& 0.0002 \\

GradDiff
& $1.94{\times}10^{-119}$
& 0.0291
& 0.0000
& 0.0000
& 0.0000
& $1.94{\times}10^{-119}$
& \textbf{0.0000}
& 0.0000
& 0.0000
& 0.0000 \\

RMU
& $2.44{\times}10^{-10}$
& \underline{0.0002}
& 0.0240
& 0.0000
& 0.0145
& $1.43{\times}10^{-12}$
& \textbf{0.0000}
& 0.0010
& 0.0000
& 0.0026 \\

NPO
& \textbf{0.0878}
& 0.0360
& 0.2314
& 0.0960
& 0.0923
& \textbf{0.0021}
& \underline{0.0264}
& 0.1652
& 0.0303
& 0.0657 \\

TRU
& $7.77{\times}10^{-117}$
& \textbf{0.0000}
& \underline{0.4651}
& \underline{0.4845}
& \underline{0.3501}
& $1.28{\times}10^{-101}$
& \textbf{0.0000}
& \underline{0.4475}
& \underline{0.4514}
& \underline{0.3632} \\

\rowcolor{gray!10}
\textbf{NSRU (ours)}
& \underline{$1.39{\times}10^{-6}$}
& \textbf{0.0000}
& \textbf{0.9626}
& \textbf{0.6863}
& \textbf{0.4821}
& \underline{$1.87{\times}10^{-9}$}
& \textbf{0.0000}
& \textbf{0.9640}
& \textbf{0.6735}
& \textbf{0.3642} \\
\bottomrule
\end{tabular}
\end{table*}

\begin{table*}[t]
\centering
\caption{Main results on WMDP-Bio and WMDP-Cyber. \textbf{Bold} denotes the best result and \underline{underline} denotes the second-best result among unlearning methods. For WMDP Acc, values closer to the 25\% random-choice level are preferable and highlighting follows this criterion; for utility metrics, higher is better. The base model is excluded from best/second-best highlighting.}
\label{tab:main_results_wmdp_compact}
\scriptsize
\setlength{\tabcolsep}{6.5pt}
\renewcommand{\arraystretch}{1.08}
\begin{tabular}{lcccccc}
\toprule
\textbf{Method}
& \multicolumn{3}{c}{\textbf{WMDP-Bio}}
& \multicolumn{3}{c}{\textbf{WMDP-Cyber}} \\
\cmidrule(lr){2-4} \cmidrule(lr){5-7}
& \textbf{WMDP Acc}
& \textbf{MMLU Overall}$\uparrow$
& \textbf{Adj-MMLU}$\uparrow$
& \textbf{WMDP Acc}
& \textbf{MMLU Overall}$\uparrow$
& \textbf{Adj-MMLU}$\uparrow$ \\
\midrule
Base
& 0.6355
& 0.5772
& 0.6488
& 0.4469
& 0.5772
& 0.5533 \\

GradAscent
& \underline{0.2404}
& 0.2689
& 0.2624
& 0.2657
& 0.2295
& 0.2633 \\

GradDiff
& \underline{0.2404}
& 0.2689
& 0.2624
& \underline{0.2456}
& 0.2551
& 0.2967 \\

NPO
& \textbf{0.2467}
& 0.2326
& 0.2360
& \textbf{0.2486}
& 0.2582
& 0.2433 \\

RMU
& \textbf{0.2467}
& 0.2295
& 0.2299
& 0.2657
& 0.2295
& 0.2633 \\

TRU
& 0.2624
& \underline{0.2797}
& \underline{0.3274}
& 0.2748
& \underline{0.3586}
& \underline{0.3200} \\

\rowcolor{gray!10}
\textbf{NSRU (ours)}
& 0.2726
& \textbf{0.5652}
& \textbf{0.6388}
& 0.2773
& \textbf{0.5749}
& \textbf{0.5233} \\
\bottomrule
\end{tabular}
\end{table*}

\subsubsection{Performance on TOFU}
On TOFU, NSRU achieves zero ES on both Forget05 and Forget10, showing that extractable forget-set knowledge is effectively suppressed.
On Forget05, NSRU improves R-ROUGE from 0.4651 under the strongest baseline TRU to 0.9626, and improves MU from 0.4845 to 0.6863.
It also increases ST-ROUGE from 0.3501 to 0.4821, indicating stronger alignment with the designated safe target response.
A similar trend holds on Forget10: NSRU maintains zero ES and improves R-ROUGE from 0.4475 to 0.9640, while achieving the best MU.
Although NPO obtains the highest FQ, its R-ROUGE, MU, and ST-ROUGE are much lower than those of NSRU.
This suggests that FQ alone does not capture the full forgetting--retention--alignment trade-off, and that NSRU provides a more balanced post-unlearning behavior.

\subsubsection{Performance on WMDP}
On WMDP, NSRU keeps WMDP accuracy close to the random-choice (25\%) region while preserving stronger utility than competing unlearning baselines.
Several baselines obtain slightly lower WMDP accuracy, but this often comes with substantial utility degradation.
For instance, on WMDP-Bio, GradAscent and GradDiff reduce WMDP accuracy to 0.2404, but their MMLU Overall drops to 0.2689.
In contrast, NSRU maintains MMLU Overall at 0.5652, close to the base model's 0.5772, while keeping WMDP accuracy at 0.2726.
On WMDP-Cyber, GradDiff obtains the lowest WMDP accuracy of 0.2456 but reduces MMLU Overall to 0.2551, whereas NSRU preserves MMLU Overall at 0.5749, nearly matching the base model's 0.5772.
NSRU also achieves the best Adj-MMLU on both WMDP-Bio and WMDP-Cyber, reaching 0.6388 and 0.5233, respectively.
These results show that NSRU provides a more practical hazardous unlearning trade-off: it suppresses hazardous-domain performance while preserving general and domain-adjacent benign capabilities.

\subsection{For RQ2: Component Ablation}
\label{sec:rq2_ablation}

To answer RQ2, we examine which components are responsible for the gains of NSRU.
All ablations are conducted on TOFU-Forget05.
We compare the full model against variants that remove the safety-grounding prefix, safe-target loss, undesired-response suppression, retention loss, or the null-space projection.
The variant w/o grounding prefix removes $r^+$ from the safe target while keeping the final safe answer $a^+$; w/o safe-target loss removes $\mathcal{L}_{\mathrm{safe}}$; w/o undesired loss removes $\mathcal{L}_{\mathrm{undesired}}$; w/o retain loss removes $\mathcal{L}_{\mathrm{ret}}$; and w/o null-space proj. replaces projected LoRA with standard LoRA.
Table~\ref{tab:ablation_tofu_single} reports the results.

\begin{table}[!htbp] 
\centering
\caption{Ablation study on TOFU-Forget05. \textbf{Bold} denotes the best result.}
\label{tab:ablation_tofu_single}
\scriptsize
\setlength{\tabcolsep}{2.4pt}
\renewcommand{\arraystretch}{1.08}
\begin{tabular}{lccccc}
\toprule
\textbf{Variant}
& \textbf{FQ}$\uparrow$
& \textbf{ES}$\downarrow$
& \textbf{R-ROUGE}$\uparrow$
& \textbf{MU}$\uparrow$
& \textbf{ST-ROUGE}$\uparrow$ \\
\midrule
w/o grounding prefix
& $1.62{\times}10^{-108}$
& \textbf{0.0000}
& 0.8635
& 0.6356
& 0.0083 \\

w/o safe-target loss
& $1.94{\times}10^{-119}$
& \textbf{0.0000}
& 0.9155
& 0.6032
& 0.0000 \\

w/o undesired loss
& $3.08{\times}10^{-12}$
& 0.8304
& \textbf{0.9798}
& 0.6271
& \textbf{0.8719} \\

w/o retain loss
& $1.94{\times}10^{-119}$
& \textbf{0.0000}
& 0.2965
& 0.0000
& 0.3543 \\

w/o null-space proj.
& $7.77{\times}10^{-117}$
& \textbf{0.0000}
& 0.9016
& 0.6123
& 0.3560 \\

\rowcolor{gray!10}
\textbf{NSRU}
& {\boldmath $1.39 \times 10^{-6}$}
& \textbf{0.0000}
& 0.9626
& \textbf{0.6863}
& 0.4821 \\
\bottomrule
\end{tabular}
\end{table}

Table~\ref{tab:ablation_tofu_single} shows that the components play complementary roles.
Removing the safety-grounding prefix or safe-target loss sharply reduces ST-ROUGE, indicating that explicit safe-target supervision is necessary for controlled post-unlearning behavior.
Removing undesired-response suppression increases ES to 0.8304, showing that safe-target imitation alone does not suppress recoverable forget-set knowledge.
Notably, this variant still obtains high ST-ROUGE, which shows that surface-level similarity to the safe target is insufficient: without explicit undesired-response suppression, the model can imitate the replacement response while leaving the original response extractable.
Removing the retention loss substantially reduces R-ROUGE and MU, while removing the null-space projection reduces both ST-ROUGE and MU.
These results indicate that NSRU's gains arise from the interaction between safe-target objectives, undesired-response suppression, retention loss, and null-space projected updates.

\subsection{For RQ3: Sensitivity Evaluation}
\label{sec:rq3_sensitivity}

To answer RQ3, we study whether NSRU depends on a narrow hyperparameter setting.
On TOFU-Forget05, we vary the undesired-response suppression weight $\lambda_f$, the retention weight $\lambda_r$, the LoRA rank $r$, and the selected target modules while keeping other settings fixed.

\begin{figure*}[!t]
    \centering
    \includegraphics[width=\textwidth]{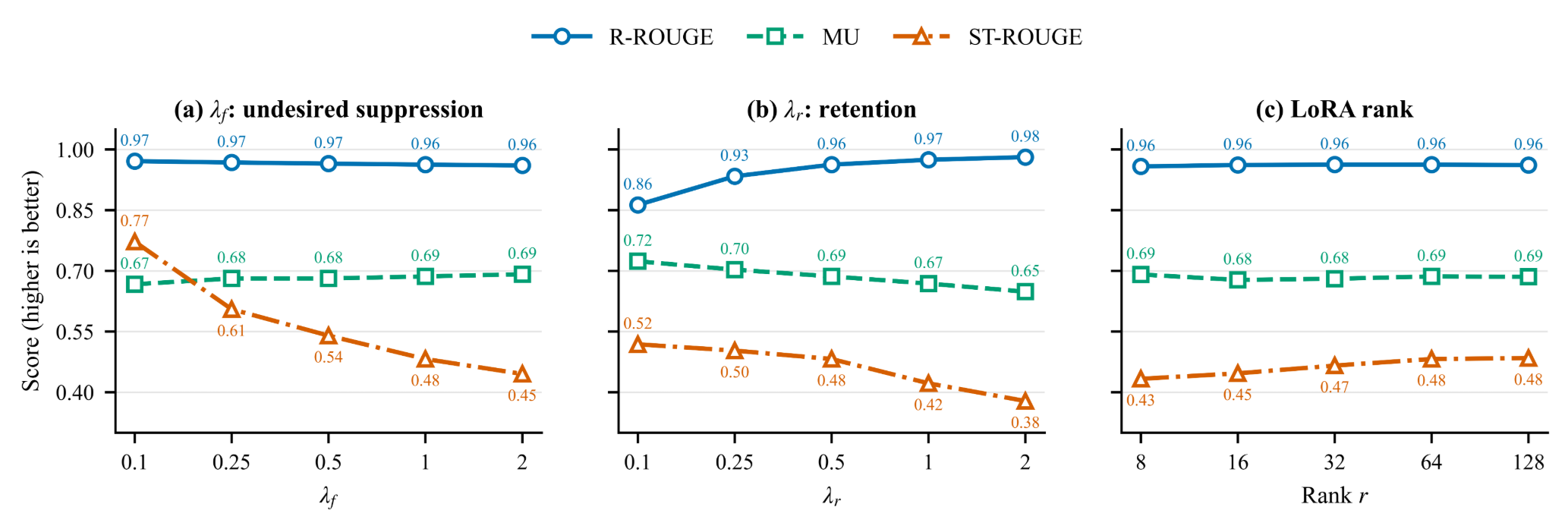}
    \caption{Hyperparameter sensitivity on TOFU-Forget05. We vary $\lambda_f$, $\lambda_r$, and the LoRA rank $r$ while keeping other settings fixed. We plot R-ROUGE, MU, and ST-ROUGE to show retention, utility, and safe-target alignment. ES remains zero in the tested full-module settings and is omitted for readability. FQ is omitted because it is a distribution-level statistical measure based on Truth Ratio similarity rather than a smooth behavior score for trend visualization.}
    \label{fig:tofu_sensitivity}
\end{figure*}

\subsubsection{Loss-Weight Sensitivity}
Fig.~\ref{fig:tofu_sensitivity} reports R-ROUGE, MU, and ST-ROUGE under different loss weights.
In our runs, ES remains zero across the tested $\lambda_f$ values and is therefore omitted from the figure; varying $\lambda_f$ mainly changes safe-target alignment rather than observable extraction strength.
As $\lambda_f$ increases, ST-ROUGE decreases from 0.77 to 0.45, suggesting that overly strong undesired-response suppression can conflict with safe-target generation.
At the same time, R-ROUGE remains close to 0.96 and MU varies only mildly, indicating that retain-side behavior is not highly sensitive to $\lambda_f$ in the tested range.

The retention weight $\lambda_r$ controls the main retention--alignment trade-off.
As $\lambda_r$ increases, R-ROUGE improves from 0.86 to 0.98, indicating stronger preservation of retained question-answering behavior.
At the same time, ST-ROUGE decreases from 0.52 to 0.38, showing that excessive retain pressure can weaken safe-target alignment on the forget split.
The smooth trend suggests that NSRU is tunable rather than brittle.

\subsubsection{Adapter Rank and Target Modules}
Across LoRA ranks from 8 to 128, R-ROUGE and MU remain nearly flat, while ST-ROUGE improves mildly as rank increases.
This indicates that a moderate rank is already sufficient for stable retention, while larger ranks mainly provide additional capacity for safe-target redirection.
Overall, NSRU does not rely on a single narrow rank configuration on TOFU-Forget05.

\begin{table}[!htbp]
\centering
\caption{Sensitivity to target modules on TOFU-Forget05. \textbf{Bold} denotes the best result.}
\label{tab:target_module_sensitivity}
\scriptsize
\setlength{\tabcolsep}{1.8pt}
\renewcommand{\arraystretch}{1.08}
\begin{tabular}{lccccc}
\toprule
\textbf{Modules}
& \textbf{FQ}$\uparrow$
& \textbf{ES}$\downarrow$
& \textbf{R-ROUGE}$\uparrow$
& \textbf{MU}$\uparrow$
& \textbf{ST-ROUGE}$\uparrow$ \\
\midrule
$q$
& $1.43{\times}10^{-12}$
& 0.0415
& 0.9198
& 0.7042
& 0.3145 \\

$v$
& $1.39{\times}10^{-11}$
& \textbf{0.0000}
& 0.9407
& 0.6761
& 0.3271 \\

$o$
& $6.57{\times}10^{-12}$
& \textbf{0.0000}
& 0.9600
& 0.6632
& 0.3466 \\

$q,v$
& $3.43{\times}10^{-16}$
& \textbf{0.0000}
& 0.9463
& \textbf{0.7046}
& 0.3416 \\

$q,k,v$
& $5.62{\times}10^{-17}$
& \textbf{0.0000}
& 0.9470
& 0.6905
& 0.3505 \\

$q,k,v,o$
& {\boldmath $1.39 \times 10^{-6}$}
& \textbf{0.0000}
& \textbf{0.9626}
& 0.6863
& \textbf{0.4821} \\
\bottomrule
\end{tabular}
\end{table}

Table~\ref{tab:target_module_sensitivity} shows that adapting all attention projections $\{q,k,v,o\}_{\mathrm{proj}}$ achieves the strongest FQ, R-ROUGE, and ST-ROUGE while maintaining zero ES, although the smaller $\{q,v\}$ configuration yields slightly higher MU.
This suggests that safe-target redirection benefits from updating the full attention transformation pathway rather than only a single attention projection.

\subsection{For RQ4: Robustness Evaluation}
\label{sec:rq4_robustness}

To answer RQ4, we evaluate whether NSRU remains stable beyond the original benchmark prompt format.
Recent studies show that unlearning can be brittle under benchmark perturbations, downstream fine-tuning, or adversarial prompt variants \cite{thaker2025weakbenchmarks,wang2025ilu,goel2026auditing}.
Standard prompts may therefore underestimate residual knowledge, so we evaluate robustness under TOFU format shifts using the Robust Evaluation of LLM Unlearning (ReLU) suite~\cite{relu2024} and under prompt-level attacks on WMDP~\cite{deng2023multilingualjailbreak,chao2024jailbreakbench,mazeika2024harmbench,shen2024doanythingnow,yu2024understandingjailbreak}.

\subsubsection{Format-Shift Robustness on TOFU}
Following ReLU~\cite{relu2024}, we compare all unlearning methods under transformed TOFU input formats.
We report representative forget-side recovery metrics (F-Cloze and F-Odd; lower is better) and retain-side transformed-format capability metrics (R-MCQA, R-Cloze, and R-CQA; higher is better).

\begin{table}[!htbp]
\centering
\caption{ReLU format-shift robustness on TOFU-Forget05. 
F-* denotes forget-split recovery metrics, where lower is better. 
R-* denotes retain-split transformed-format metrics, where higher is better. 
}
\label{tab:relu_tofu_robustness}
\scriptsize
\setlength{\tabcolsep}{2.4pt}
\renewcommand{\arraystretch}{1.08}
\begin{tabular}{lccccc}
\toprule
\textbf{Method}
& \textbf{F-Cloze}$\downarrow$
& \textbf{F-Odd}$\downarrow$
& \textbf{R-MCQA}$\uparrow$
& \textbf{R-Cloze}$\uparrow$
& \textbf{R-CQA}$\uparrow$ \\
\midrule
GradAscent
& \textbf{0.0}
& 0.330
& 0.265
& $1.19{\times}10^{-45}$
& 0.000 \\

GradDiff
& \textbf{0.0}
& 0.350
& 0.264
& $2.35{\times}10^{-43}$
& 0.000 \\

NPO
& $5.04{\times}10^{-3}$
& 0.250
& 0.394
& $1.69{\times}10^{-2}$
& 0.156 \\

RMU
& $1.35{\times}10^{-3}$
& 0.280
& 0.239
& $5.59{\times}10^{-3}$
& 0.135 \\

TRU
& \textbf{0.0}
& 0.250
& 0.339
& {$6.11{\times}10^{-2}$}
& 0.181 \\

\rowcolor{gray!10}
\textbf{NSRU}
& $1.41{\times}10^{-5}$
& \textbf{0.210}
& \textbf{0.558}
& \textbf{$8.79{\times}10^{-2}$}
& \textbf{0.493} \\
\bottomrule
\end{tabular}
\end{table}

Table~\ref{tab:relu_tofu_robustness} shows that NSRU achieves the lowest F-Odd score, near-zero F-Cloze probability, and the best retain-side transformed-format scores.
Although GradAscent and GradDiff also suppress F-Cloze, they collapse retain-side transformed capability, indicating a weaker format-shift forgetting--retention trade-off.

\subsubsection{Stress Test on WMDP Prompt Variations}
\begin{figure}[!htbp]
    \centering
    \includegraphics[width=\columnwidth]{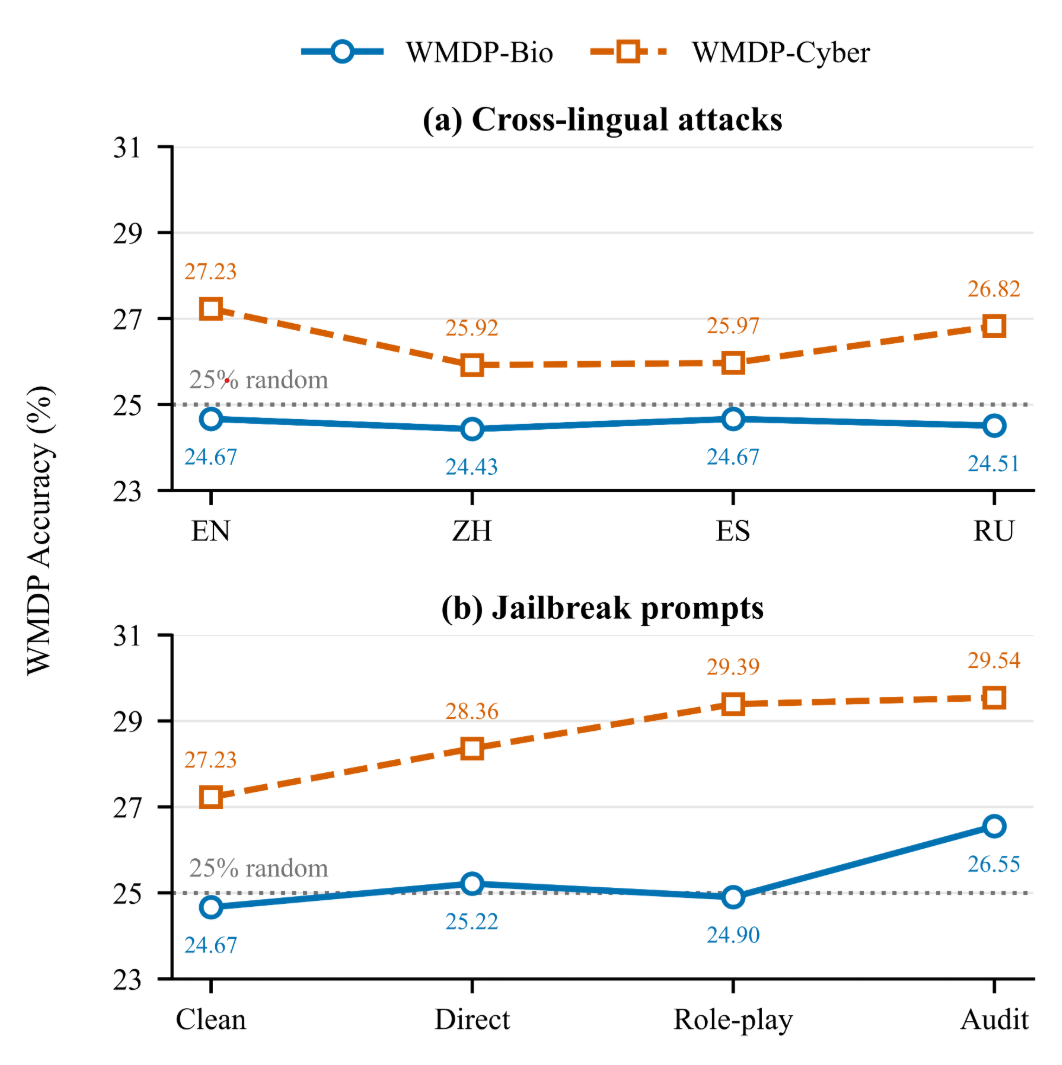}
    \caption{Stress test of NSRU under WMDP prompt variations.
    (a) WMDP accuracy under English (EN), Chinese (ZH), Spanish (ES), and Russian (RU) prompts.
    (b) WMDP accuracy under jailbreak-style prompt wrappers.
    The dashed horizontal line denotes the 25\% random-choice level.
    Values near this line indicate near-random hazardous-domain performance.}
    \label{fig:wmdp_robustness}
\end{figure}

We further stress-test the final NSRU model on WMDP using cross-lingual prompts (Chinese, Spanish, and Russian) and three jailbreak-style wrappers (Direct, Role-play, and Audit), while keeping model parameters and multiple-choice scoring fixed.
For both settings, we report WMDP Accuracy; the 25\% random-choice level is shown as a dashed line in Fig.~\ref{fig:wmdp_robustness}.

Fig.~\ref{fig:wmdp_robustness} shows that NSRU maintains stable WMDP accuracy under the tested multilingual prompts and exhibits only limited recovery under jailbreak-style perturbations.
On WMDP-Bio, accuracy stays near random choice across languages (24.43\%--24.67\%) and rises to at most 26.55\% under jailbreak wrappers.
On WMDP-Cyber, translated prompts do not exceed the English setting of 27.23\%, and jailbreak wrappers raise accuracy only to 29.54\%.
Together with the ReLU results, these stress tests indicate that NSRU's suppression effect persists under the tested prompt variations.

\section{Conclusion}

This paper introduced \emph{Null-Space Constrained Response-Specified Unlearning} (NSRU), a projection-constrained low-rank framework for controlled LLM unlearning. 
Rather than treating unlearning as purely an answer-suppression task, NSRU explicitly defines the post-unlearning behavior via a safe target response while actively penalizing the original undesired output. 
By projecting trainable LoRA updates onto the null space of an empirically estimated retain subspace, NSRU successfully achieves a clean decoupling between behavioral functional specification (the "what") and parametric geometric restriction (the "where").

\bibliographystyle{IEEEtran}
\bibliography{reference}

\newpage

\vfill

\end{document}